\documentclass{article}

\usepackage[preprint]{neurips_2023}

\usepackage[utf8]{inputenc} 
\usepackage[T1]{fontenc}    
\usepackage{hyperref}       
\usepackage{url}            
\usepackage{booktabs}       
\usepackage{amsfonts}       
\usepackage{nicefrac}       
\usepackage{microtype}      
\usepackage{xcolor}         

\usepackage{epsfig}
\usepackage{graphicx}
\usepackage{amsmath}
\usepackage{amssymb}
\usepackage{amsthm}
\usepackage{amsmath}
\usepackage{bm}
\usepackage{enumerate}
\usepackage{algorithm2e}
\usepackage{thm-restate}
\usepackage{caption}
\usepackage{subcaption}
\usepackage{float}
\usepackage{multirow}
\usepackage{multicol}
\usepackage{times}
\usepackage{sidecap}
\usepackage{wrapfig,lipsum,booktabs}

\title{Variational Inference for Scalable 3D Object-centric Learning}

\author{%
  Tianyu Wang \\
  School of Computing\\
  The Australia National University\\
  \texttt{tianyu.wang2@anu.edu.au}\\
  \And
  Kee Siong Ng\\
  School of Computing\\
  The Australia National University\\
  \texttt{keesiong.ng@anu.edu.au}\\
  \And
  Miaomiao Liu\\
  School of Computing\\
  The Australian National University\\
  \texttt{miaomiao.liu@anu.edu.au}\\
}

\begin{document}

\maketitle

\begin{abstract}
  We tackle the task of scalable unsupervised object-centric representation learning on 3D scenes. 
  Existing approaches to object-centric representation learning show limitations in generalizing to larger scenes as their learning processes rely on a fixed global coordinate system. 
  In contrast, we propose to learn view-invariant 3D object representations in localized \emph{object coordinate systems}. 
  To this end, we estimate the object pose and appearance representation separately and explicitly map object representations across views while maintaining object identities. 
  We adopt an amortized variational inference pipeline that can process sequential input and scalably update object latent distributions online. 
  To handle large-scale scenes with a varying number of objects, we further introduce a \emph{Cognitive Map} that allows the registration and query of objects on a per-scene global map to achieve scalable representation learning. 
  We explore the object-centric neural radiance field (NeRF) as our 3D scene representation, which is jointly modelled within our unsupervised object-centric learning framework. 
  Experimental results on synthetic and real datasets show that our proposed method can infer and maintain object-centric representations of 3D scenes and outperforms previous models. 
\end{abstract}

\section{Introduction}
  In recent years, 2D and 3D unsupervised object-centric learning has attracted increasing attention. 
  While 2D object-centric learning methods \citep{AIR, SPACE, MONET, SPAIR, SlotAttention} aim to identify and segment objects within images, 3D methods aim to reconstruct complete 3D scene structures in an object-centric manner using RGB or RGBD observations \citep{MulMON, ObSuRF, ROOTS, ObjectCentricVideoGeneration}. 
  The ability to understand 3D surroundings in an object-centric way is crucial for high-level tasks such as relational reasoning and object manipulation.
  The majority of existing 3D methods assume that target scene scales are small enough to fit into the field of view (FOV) of a single camera and are centered at the origin of pre-defined global coordinate systems \citep{MulMON, ObSuRF, ROOTS}. 
  As a result, these models fail to generalize to scenes beyond training set scale.

  In this work, we aim to remove the small-scene assumption and to handle scene of potentially unbounded scales.
  As each camera view can only capture a limited local region of a scene, obtaining a comprehensive scene representation requires aggregating information from a potentially unknown number of diverse views. To this end, we propose as solution a 3D object-centric learning pipeline termed \textit{\textbf{S}calable \textbf{O}nline \textbf{O}bject \textbf{C}entric network in {\bf 3D} (SOOC3D)}.
  Specifically, SOOC3D formulates object-centric learning as an online latent variable inference problem, explicitly models object poses and infers view-invariant object representations in localized ~\emph{object~coordinate~systems}.
  To maintain object identities during the inference process, we introduce a scalable memory mechanism named~\emph{Cognitive Map},\footnote{The term cognitive map is borrowed from cognitive psychology studies on mental representations of the spatial surroundings in animal, and human brain \citep{CognitiveMap}.} which can be used to register and query detected objects.
  Our proposed model is an unsupervised learning framework that is trained to reconstruct RGBD observations using  object-compositional neural radiance field (NeRF).
  Previous works show that object-centric NeRF models commonly exhibit lower reconstruction quality compared to per-scene NeRFs optimised directly with SGD \citep{NeRF, zhang2022nerfusion, tancik2022block}. This is due to the network bottlenecks filtering out high-frequency information \citep{OCIB}.
  We show that our framework supports a scalable per-object NeRF finetuning process which improves the reconstruction quality with preserved object identities.

    Our contributions are summarised as follows. 
    i) We propose, to the best of our knowledge, the first unbounded scalable generative-model-based unsupervised 3D object-centric learning framework.
    ii) We learn the explicit object locations and view-invariant object representations separately via the amortized variational inference framework to achieve scalable online updating. 
    iii) To store a potentially unbounded number of detected objects for scalable inference, we introduce \emph{Cognitive Map} separating object representations management from the inference process.
    iv) We demonstrate that the reconstruction quality can be further improved via our per-object NeRF finetuning process.

    \begin{figure*}[h]
      \hfill
      \begin{subfigure}[c]{0.45\linewidth}
          \centering
          \includegraphics[width=\textwidth]{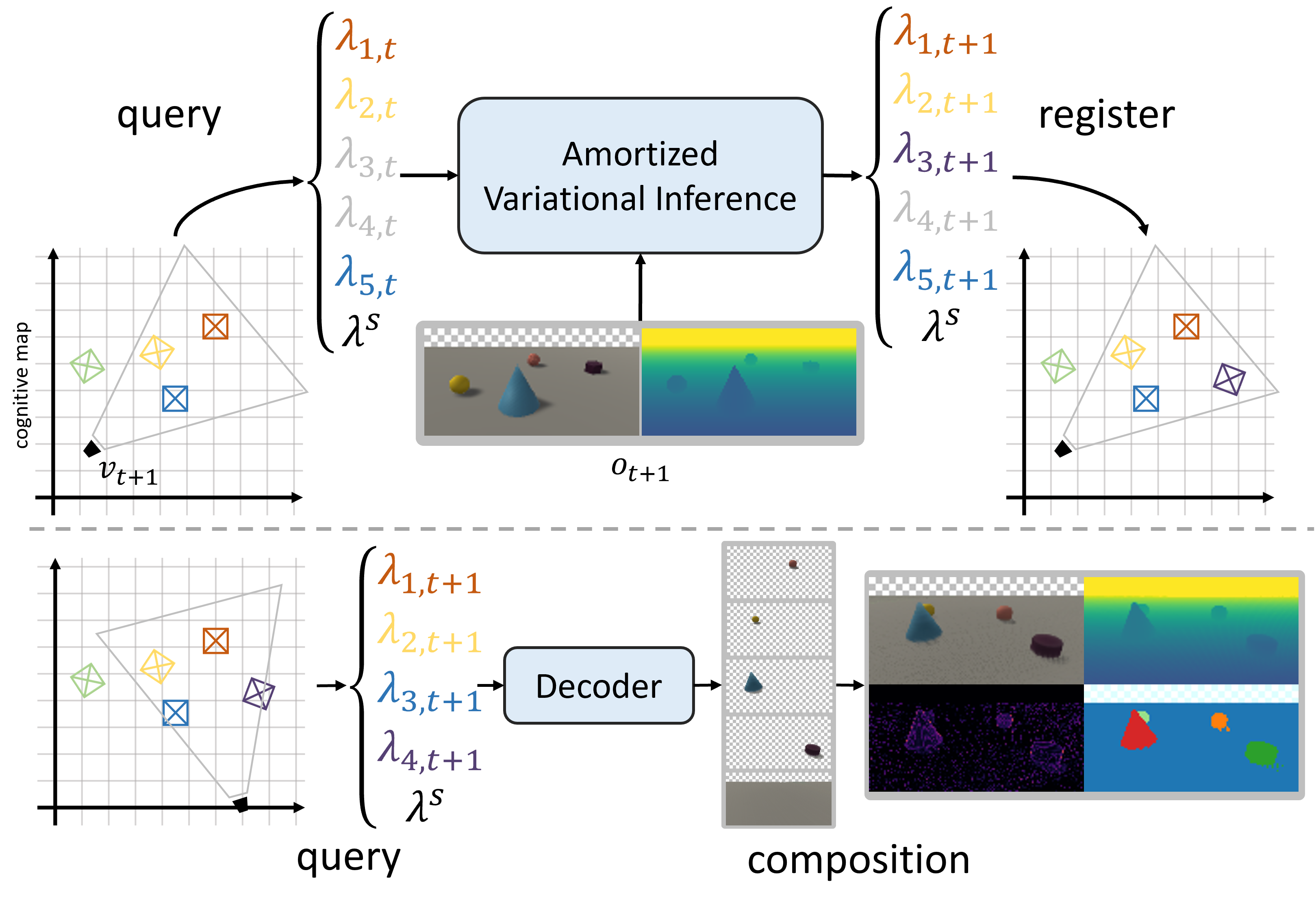}
      \end{subfigure}
      \hfill
      \begin{subfigure}[c]{0.45\linewidth}
          \centering
          \includegraphics[width=\textwidth]{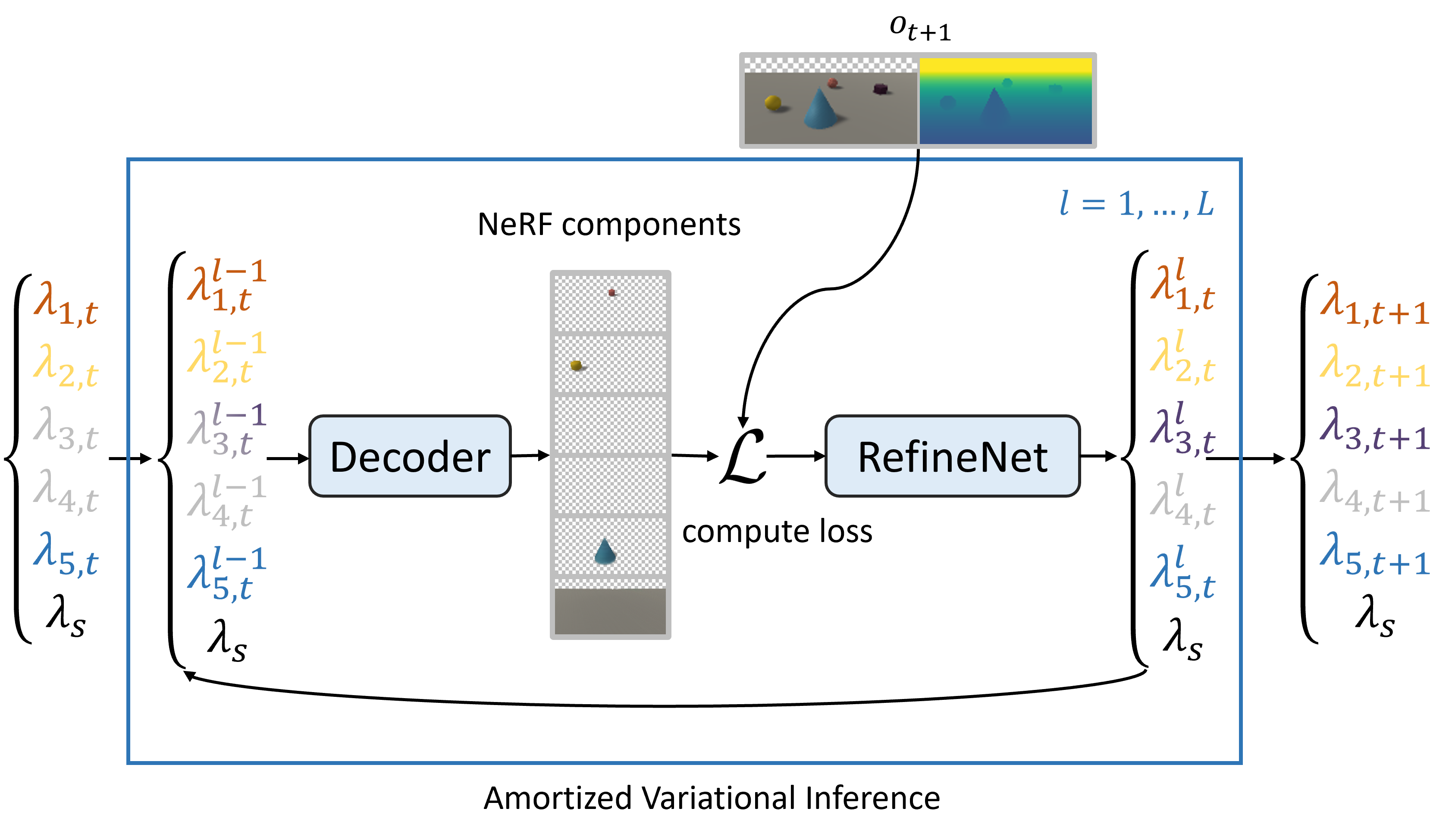}
      \end{subfigure}
      \hfill
          \caption{\textbf{Left}: The online inference process for scene updating (top) and novel view synthesis (bottom).
          Previously detected objects are registered in the cognitive map.
          Given a new observation, the representations of all objects in FOV are retrieved. 
          Distributions over object latent variables and scene layout latent variables are parameterized by \{$\lambda_{i, t}$\} and $\lambda^s$.
          If the number of objects existing in the current view is less than a pre-defined value $K$, we pad with priors (greyed $\lambda_{i, t}$).
          Amortized variational inference process updates $\{\lambda_{i, t}\}$ to integrate new information.
          Finally, we register updated $\{\lambda_{i, t+1}\}$ back into the cognitive map.
          For novel view synthesis, we sample latent variables and decode them into NeRF components.
          By composition, we obtain RGB, depth, segmentation and uncertainty map.
          \textbf{Right}: An $L$-iteration amortized variational inference process.
          In each iteration, the set of input representations is decoded into NeRF.
          The refinement network takes raw observation, reconstruction and other auxiliary variables to update latent variables.
          }
          \label{fig:net}
  \end{figure*}

\section{Related Work}

  \textbf{Unsupervised 2D Object-centric Learning.}
  2D object-centric learning aims to group pixels covering the same object under the same label and at the same time produce a neural representation of each discovered object. 
  At the core of those methods is the spatial mixture model formulation that frames object-centric learning as a latent variable inference problem \citep{AIR, IODINE, NEM}.
  To handle observations of high object density, a branch of works~\citep{AIR, SPAIR, SPACE} infers latent variables for local regions of each 2D observation. 
  Pipelines equipped with iterative refinement modules \citep{NEM, SlotAttention} refine the latent variable iteratively conditioned on an input view.  
  Particularly, IODINE \citep{IODINE} employs amortized variational inference \citep{IAI} that can process sequential data. 
  In dynamic scenes, motion cues can be exploited to improve the segmentation results \citep{STEVE,PPMP}.
  However, the aforementioned methods do not infer 3D structures.
  Object latents are discarded once out of view.

  \textbf{Unsupervised 3D Object-centric Learning.}
  3D-aware methods not only try to factorize observations in an object-centric manner but also infer the spatial structure of scenes, which can be examined by the means of novel views synthesis.
  Similar to its 2D counterpart, 3D object-centric representation learning approaches also adopt the spatial mixture model formulation.
  ObSuRF~\citep{ObSuRF} and uORF~\citep{uORF} introduce neural radiance fields (NeRF) into the object-centric learning setting.
  \cite{ObjectLightField} propose to use object light field to avoid dense sampling along rays during rendering.
  As an attempt to model scenes on a larger scale, O3V~\citep{ObjectCentricVideoGeneration} and SIMONe~\citep{SIMONe} infer object-centric representation from a video sequence. 
  However, both O3V and SIMONe adopt a non-incremental method and process entire video sequences before generating scene representations.
  MulMON~\citep{MulMON} adopts amortized variational inference to allow object latents to be updated by new views online.
  The methods mentioned above work well on scenes with sizes that can fit into the camera FOV but fail to scale up to larger scenes.

  \textbf{Object-Compositional NeRF and Scalable NeRF.}
  Object-compositional NeRF has been studied recently to learn the 3D representation of each object and the scene for image synthesis.
  In particular, \cite{yang2021learning} introduced a two-pathway framework to model the foreground objects and the scene branch, with known coarse object instance masks.
  Such a method cannot be directly applied to the unsupervised scenario. 
  To handle large scenes, block-wise NeRF has been proposed recently\citep{zhang2022nerfusion,tancik2022block}. 
  They can either generalize to large scene \citep{zhang2022nerfusion} by taking multiple images as input or train scene-specific block-wise NeRFs for large scene fast rendering~\citep{zhang2022nerfusion}. 
  However, both~\cite{zhang2022nerfusion} and \cite{tancik2022block} are not object aware. 
  In this paper, we aim to combine the merits from all the approaches and learn scalable object-centric NeRF scene representations.

\section{Method}
  We formulate object-centric representation learning as a latent variable inference problem with a dynamic latent variable set (Sec. \ref{sec:GenerativeModel} (a)).
  We then present a factorized variational proposal distribution tailored for scalable online inference (Sec. \ref{sec:AVI} (b)).
  Finally, we present implementation details of the proposed inference pipeline (Sec. \ref{sec:ModelStructure}). 

  \subsection{Formulation and the Optimization Objective}
  \textbf{(a) Generative Model.}
  \label{sec:GenerativeModel}
  At each time $t$ a camera of known pose $v_t$ captures an RGB-D frame $o_{v_t}$ of the scene. 
  $v_t$ is the extrinsic parameters of a camera in an arbitrary global coordinate system. 
  Under the static scene assumption, each camera induces a fixed frame. 
  Note that we do not pre-set an end time for data receiving, thus, $t$ can be unboundedly large.
  To describe the observation received up to time t, we denote $\mathcal{O}_t$ = $\{o_{v_1}, \dots, o_{v_t}\}$ and $\mathcal{V}_t = \{v_1, \dots, v_t\}$.
  We assume the entire scene is described by a set of latent variables $\mathcal{Z} = \{z_1, z_2, \dots\}$ of potentially unbounded size. 


  The exact latent variable posterior $p(\mathcal{Z} | \mathcal{O}_t, \mathcal{V}_t)$ is intractable.
  Considering that $t$ is not assumed to be bounded, it is infeasible to process all views at once and an online approximate inference method is needed.
  Thus, we resort to the amortized variational inference framework \citep{IODINE, MulMON, EIAI}.

  \textbf{(b) Variational Inference.}
  \label{sec:AVI}

  We denote the latent variable set after observing $\mathcal{O}_t$ and $\mathcal{V}_t$ as $\mathcal{Z}_t = \{z_{1,t}, \dots z_{m_t, t}\}$ and decompose latent variable $z_{i,t}$ as $z_{i,t} = (z^{where}_{i,t}, z^{what}_{i,t})$.
  $z^{what}_{i, t} \in \mathbb{R}^d$ is the object appearance embedding.
  While object poses in general have 6 degrees of freedom (DoF), under static scene assumption, 3 DoF is sufficient to achieve scalable inference.
  Thus, we only model its location on $xz$-plane and rotation about $y$-axis and set $z^{where}_{i, t} \in \mathbb{R}^3$.
  The other 3 DoF are modelled within object appearance embeddings implicitly.

  Below, we show a way to approximate the exact posterior $p(\mathcal{Z} | \mathcal{O}_t, \mathcal{V}_t)$ with a parameterized proposal distribution $q(\mathcal{Z}_{t} | \mathcal{O}_t, \mathcal{V}_t)$.
  By exploiting the temporal and spatial structure of the problem, we simplify $q(\mathcal{Z}_{t} | \mathcal{O}_t, \mathcal{V}_t)$ to allow online and scalable inference.

  \textbf{Temporal factorization:}
  First, we exploit the temporal structure of the problem and recursively factorize the proposal distribution as
  \begin{equation}
      \label{eq:temporal}
      q(\mathcal{Z}_{t} | \mathcal{O}_t, \mathcal{V}_t) = \int q(\mathcal{Z}_{t} | o_{v_t}, v_t, \mathcal{Z}_{t-1}) q(\mathcal{Z}_{t-1} |\mathcal{O}_{t-1}, \mathcal{V}_{t-1}) d \mathcal{Z}_{t-1},
  \end{equation}
  where $q(\mathcal{Z}_{t-1} |\mathcal{O}_{t-1}, \mathcal{V}_{t-1})$ is the posterior from the previous step and $q(\mathcal{Z}_{t} | o_{v_t}, v_t, \mathcal{Z}_{t-1})$ is the update distribution.
  This recursive factorization greatly simplifies the computation and allows us to update the posterior at each step by computing $q(\mathcal{Z}_{t} | o_{v_t}, v_t, \mathcal{Z}_{t-1})$ for unbounded $t$.

  \textbf{Spatial factorization:}
  With $z^{where}_{i, t}$ explicitly modeled, we can exclude from the update distribution the latent variables that are out of the FOV of camera $v_t$ by assuming that such latent variables will not contribute to the observation generation.
  For a latent set $\mathcal{Z}_{t'}$, we denote the set of latents in FOV of view $v_{t}$ as $\mathcal{Z}_{t'}^{v_t}$ and its complement as $\bar{\mathcal{Z}}_{t'}^{v_t} = \mathcal{Z}_{t'} \backslash \mathcal{Z}_{t'}^{v_t}$ (the set of out-of-view latents).
  This spatial structure allows us to factorize the update distribution as
  \begin{equation}
    \label{eq:spatial}
      q(\mathcal{Z}_{t} | o_{v_t}, v_t, \mathcal{Z}_{t-1}) = q(\mathcal{Z}_{t}^{v_t} | o_{v_t}, v_t, \mathcal{Z}_{t-1}^{v_t})q(\bar{\mathcal{Z}}_{t}^{v_t} | o_{v_t}, v_t, \bar{\mathcal{Z}}_{t-1}^{v_t}),
  \end{equation}
  where $q(\bar{\mathcal{Z}}_{t}^{v_t} | o_{v_t}, v_t, \bar{\mathcal{Z}}_{t-1}^{v_t}) = \delta_{\bar{\mathcal{Z}}_{t-1}^{v_t}}(\bar{\mathcal{Z}}_{t}^{v_t})$ (Dirac delta).
  That is, latent variables that are not related to current observations remain unchanged.
  Now we can focus on the updated distribution of in-view latent sets $q(\mathcal{Z}_{t}^{v_t} | o_{v_t}, v_t, \mathcal{Z}_{t-1}^{v_t})$.

  \textbf{Surrogate distribution:} 
  $q(\mathcal{Z}_{t}^{v_t} | o_{v_t}, v_t, \mathcal{Z}_{t-1}^{v_t})$ is a distribution over a set.
  Both the set size and elements are random variables.
  A general solution is to generate set elements in an autoregressive manner, which is strictly sequential.
  For efficiency reason, we resort to parallelizable methods and assume that there can be at most $K$ objects in each view during training.
  This assumption allows us to adopt a surrogate update distribution and parallelize the inference. 

  We define the augmentation of a latent variable $z_{i, t}$ as $\hat{z}_{i, t} = (z_{i, t}, z_{i,t}^{pres})$ with $z_{i,t}^{pres} \in \{0, 1\}$.
   $z_{i,t}^{pres}$ represents the existence of each object with $z_{i,t}^{pres} = 0$ indicating that the object does not exist.
  The augmentation of $\mathcal{Z}_{t-1}^{v_t}$ is defined as 
  \begin{equation}
    \label{eq:tilde}
      \mathcal{\tilde{Z}}_{t-1}^{v_t} = \{\hat{z}_{i, t} | z_{i, t} \in \mathcal{Z}_{t-1}^{v_t} \} \bigcup \{\underbrace{\hat{z}_{0, 0}, \dots, \hat{z}_{0, 0}}_{K - |\mathcal{Z}_{t-1}^{v_t}|}\},
  \end{equation}
  where $\hat{z}_{0, 0} := (z_{0, 0}^{what}, z_{0, 0}^{where}, z_{0, 0}^{pres})$ is learnable global variational prior. 
  Intuitively, if there are less than $K$ elements in $\mathcal{Z}_{t-1}^{v_t}$ we pad with the global priors to $K$ elements.

  The surrogate update distribution is denoted as $q(\mathcal{\hat{Z}}_{t}^{v_t} | o_{v_t}, v_t, \mathcal{\tilde{Z}}_{t-1}^{v_t})$.
  $\mathcal{\hat{Z}}_{t-1}^{v_t}$ is a set of $K$ augmented latent variables holding the updated latent variables.
  We factorize the surrogate distribution as
  \begin{equation}
    \label{eq:update}
      q(\mathcal{\hat{Z}}_{t}^{v_t} | o_{v_t}, v_t, \mathcal{\tilde{Z}}_{t-1}^{v_t}) = \prod_{\hat{z} \in \mathcal{\hat{Z}}_{t}^{v_t}} q(\hat{z} | o_{v_t}, v_t, \mathcal{\tilde{Z}}_{t-1}^{v_t}).
  \end{equation}
  and implement it as a neural network. Finally, we recover the set of latent variables via
  \begin{equation}
    \label{eq:transform}
      \mathcal{Z}_{t}^{v_t} = \{z_{i, t} | \hat{z}_{i,t} \in \mathcal{\hat{Z}}_{t}^{v_t}, z_{i, t}^{pres} = 1 \}.
  \end{equation}

  We parameterize $q(z^{where}_{i, t})$ and $q(z^{what}_{i, t})$ as isotropic Gaussian with parameter $\lambda^{where}_{i, t} = \{\mu^{where}_{i, t}, \sigma^{where}_{i, t}\}$ and $\lambda^{what}_{i, t} = \{\mu^{what}_{i, t}, \sigma^{what}_{i, t}\}$.
  $q(z^{pres}_i)$ takes the form of Bernoulli distribution with $\lambda^{pres}_{i, t}$ being the logit.
  Continuous relaxation \cite{relaxB} is used for differentiable sampling.
  Scene layouts (floor) have modalities drastically different from objects.
  We define scene layout latent variables following the same definition but with fixed $z^{pres} = 1$ and pre-defined $z^{where}$ \cite{tancik2022block}.
  For each view, one and only one scene layout variable is active.

  Given the defined variational posterior above, we identify the core of the posterior inference to be $q(\hat{z} | o_{v_t}, v_t, \mathcal{\tilde{Z}}_{t-1}^{v_t})$ and we implement it via a refinement network $f_{\vartheta}$.
  For each input view $v_t$, we update our parametrized latent for $L$ iterations.
  At iteration $l \in \{1,\cdots,L\}$, the latent is updated as
  \begin{align}
      \label{eq:refine}
      \hat{z}_{i, t}^{l} &\sim q_{\lambda_{i, t}^l}(\hat{z}_{i, t}^{l}) \\
      \lambda_{i, t}^l &= \lambda_{i, t}^{l-1} + f_{\vartheta}(\hat{z}_{i, t}^{l-1}, o_{v_t}, v_t, \mathbf{a})
  \end{align}

  with $q_{\lambda_{i, t}^0} = q_{\lambda_{i, t-1}^L}$. 
  That is, the refinement results of previous steps serve as the prior of the next step.
  The desired posterior $q(\hat{z}_{i, t} | \mathcal{O}_t, \mathcal{V}_t)$ is parameterized by $\lambda_{i, t}^L$.
  $\mathbf{a}$ is a collection of auxiliary input.
  Latent variables are updated in parallel in each iteration.
  Specification of the auxiliary input and an algorithmic summary of the algorithm are provided in the supplementary.

  \textbf{(c) Training Objective.}
  Below we use $\lambda_{:,t}^l$ to denote the parameters of all latent variables at time $t$ after iteration $l$.
  Our training objective contains 3 terms. The input view observation log-likelihood is $\mathcal{L}^{input} = \sum_{t=1}^T \mathbb{E}_{q_{\lambda_{:,t}^L}} \left[\log p(o_{v_t} | v_t, \hat{\mathcal{Z}}^{v_t}_{t})\right]$, which is the sum of the log-likelihood of individual step.
  The KL term is $\mathcal{L}^{kl} = \sum_{t=1}^T \mathcal{D}_{KL}[q_{\lambda_{:,t}^L} || q_{\lambda_{:,t-1}^L}]$.
  The first two terms form the evidence lower bound (ELBO) of the intractable likelihood.
  To encourage the learning of view-invariant representation, in addition to the $T$ input views, we also sample a set of query views $\mathcal{Q}$ and compute the likelihood $\mathcal{L}^{query} = \sum_{(v, o_v) \in \mathcal{Q}} \mathbb{E}_{q_{\lambda_{:,t}^L}} \left[\log p(o_v | v, \hat{\mathcal{Z}}^{v}_T)\right]$.
  The training objective is 
  \begin{equation}
      \label{eq:elbo}
      \mathcal{L} = \mathcal{L}^{input} - \mathcal{L}^{kl} + \mathcal{L}^{query}
  \end{equation}
  By adopting the depth-informed NeRF likelihood function \cite{ObSuRF}, we are able to estimate the likelihood with two samples per ray.
  Details on training objectives are provided in the supplementary.

  \subsection{Model Implementation}
  \vspace{-3mm}
  The learning of view-invariant object appearance representation heavily relies on the pose-induced local coordinate system (a), within which we decode NeRF to reconstruct observation (b).
  Then we detail our refinement network implementation (c).
  The latent variable in-view test is implemented within a \emph{Cognitive Map} data structure which also keeps track of all latent variables through the inference process (d).
  Then, we describe our curriculum learning setup (e) and how our per-object finetuning pipeline can be applied on top of our inference results (f).
  Below we denote retrieved latents as $\hat{z}_{\phi_v(k)}$ where $\phi_v(k)$ is the global index of the $k^{th}$ element of $\tilde{\mathcal{Z}}^v$ and omit subscript if clear from the context.
  The model pipeline is shown in Fig. \ref{fig:net}. 

  \textbf{(a) Object Coordinate System.}
  Each $z^{where}_{\phi_v(k)}$ corresponds to a pose of an object component in the current camera coordinate system.  
  For each component, we build a matrix $\Pi(z^{where}_{\phi_v(k)}) \in SE(3)$ to map each point $x$ in the camera coordinate system to the local coordinate system $x_k = \Pi(z^{where}_{\phi_v(k)}) \cdot x$.
  Before decoding, each point is mapped into all object local coordinate systems.
  Note that $x_k$ is a differentiable function of $z^{where}_{\phi_v(k)}$.
  Thus, all gradients to $x_k$ flow through $z^{where}_{\phi_v(k)}$.


  \textbf{(b) Object-aware NeRF Decoding.}
  \label{sec:nerf}
  Conditioned on $z^{what}_{\phi_v(k)}$, a NeRF decoder assigns each point $x_k$ a raw density $\tilde{\sigma}_k(x_k) \in [0, 1]$ and a RGB value $\tilde{c}_k(x_k)$.
  Crucial to position learning, we introduce an inductive bias in the form of Gaussian weighting.
  To be more precise, we compute the weighted density $\log \hat{\sigma}_k(x_k) = \log \tilde{\sigma}_k(x_k) + \log w_g(x_k) - \mathcal{SG}(\log w_g(x_k)) + \log z_{\phi_v(k)}^{pres}$ where $w_g(\cdot)$ is a zero mean gaussian function and $\mathcal{SG}$ is the stop gradient operation.
  By adding $\log w_g(x_k)- \mathcal{SG}(\log w_g(x_k))$, we encourage the $z_{\phi_v(k)}^{where}$ to be close to object center but the value of the weighted density remain unchanged.
  Weighted by $z_{\phi_v(k)}^{pres}$, non-existent components are turned off.

  We then compute the normalized density as $\bar{\sigma}_k(x_k) = \frac{\hat{\sigma}_k(x_k)^2}{\sum_{i=0}^K \hat{\sigma}_i(x_i)}$.
  Note that $\sum_k \bar{\sigma}_k(x_k) \in [0, 1]$ allowing us to represent concrete object or void space. 
  The final NeRF density at point $x$ is given by $\sigma(x) = \sigma_{max} \cdot \sum_{k=0}^K \bar{\sigma}_k(x_k)$  with $\sigma_{max}$ being the maximum NeRF density of our choice.
  The color is given by $c(x) = \sum_{k=0}^K \frac{\hat{\sigma}_k(x_k)}{\sum_{i=0}^K \hat{\sigma}_i(x_i)} \tilde{c}_k(x_k)$.

  During training, it is sufficient to evaluate one sample on the surface and one sample in the air for each ray (see the supplementary for details).
  For testing, we densely sample the camera frustum, evaluate samples and compose via rendering equations. 

  \textbf{(c) Refinement Network.}
  \label{sec:ModelStructure}
  At each time step $t$, give observation $o_{v_t}, v_t$, the refinement network implements the update distribution defined in Eq. \ref{eq:update}.
  Following the amortized variational inference literature \citep{IODINE, MulMON}, the refinement network takes as input latent variables and a set of auxiliary data and outputs the updated distributions.
  The auxiliary data may include observation, reconstruction, likelihood and gradient to latent variables, etc.
  Crucial to scalability, the refinement network interprets $z^{where}$ as object locations in the camera coordinate system of $v_t$. 
  That is, the refinement network is global coordinates and scene scale agnostic.
  
  For each input view, the refinement process is executed in parallel for all components for $L$ times.
  While new objects may be detected and old objects are out-of-view when cameras are switched, we do not hard code any object-matching heuristics.
  Similar to NeRF decoder, all object latent variables share one refinement network while the scene layout component has its own refinement network.
  Network structure and auxiliary input specifications are presented in the supplementary.

  \textbf{(d) Cognitive Map.}
  \label{sec:CogMap}
  A cognitive map stores all latent variables, implements the in-view test and interacts with the inference process via the registration and the query process.

  \textbf{Registration:} 
  After each refinement step, we register the updated latent variables.
  Recall that refinement results in a set of augmented latent variables.
  We first discard all $\hat{z}$ with $q(z^{pres} = 1) < 0.5$ since they are deemed non-existent. 
  Given the camera pose, all $z^{where}$ are transformed from the camera coordinate system (where refinement happens) into the global coordinate system.
  Then $\hat{z}$ are stored in a list for future queries.

  \textbf{Query:} 
  Before each refinement step, we query all in-view latent variables. 
  Given a camera pose, we go through all registered latent variables and performance in-view checks.
  If less than $K$ objects are found, pad with priors.
  For the first view in a scene, only priors are returned. 
  In practice, it is possible to find more than $K$ objects in one view.
  In this case, we retrieve those with top-K $\lambda^{pres}$ value.
  Then all retrieved $z^{where}$ are transformed from the global coordinate system into the camera coordinates system for refinement.

  The cognitive map only exposes relevant latent variables to the inference process.
  Thus, the inference memory consumption is independent of the total number of latent variables.
  During training, gradients of latent variables can flow through the registration-query loop.
  During testing, cognitive maps can be deployed on any storage.

  \textbf{(f) Per-Object NeRF Finetuning.}
  \label{sec:OCNeRF}
  We implement the per-object NeRF finetuning using the expectation-maximization algorithm.
  The latent variables are now the object identities of all ray samples.
  As initialization, we duplicate the trained NeRF decoder for each latent in the cognitive map.
  During the finetuning process, $z^{where}$ is fixed and $z^{what}$ is treated as part of per-object NeRF parameters.
  The evidence lower bound is maximized via direct
  gradient descent.
  Detailed formulations are provided in the supplementary.
  The finetuning process does not diminish the scalability as the per-object NeRF can also be registered into the cognitive map for future queries or finetuning.

  \section{Experiments}
  \textbf{Dataset.}
  The datasets commonly used in object-centric learning literature are often limited in scene scale, making them unsuitable for scalability studies \citep{gqn, CLEVR, GENESIS-V2, uORF}.
  Thus, we constructed two datasets consisting of multi-view RGBD data.
  The Unity dataset is created using Unity3D \cite{Unity} and consists of simple geometry mimicking the object room dataset \cite{gqn}.
  While the Unity dataset is limited in visual complexity, it contains scenes of three different scales termed as \textit{small} (s), \textit{medium} (m) and \textit{large} (l).
  The \textit{small} scenes are similar to those in previous datasets in that they can roughly fit into the field of view of a single camera.
  The \textit{medium} scenes in this dataset are twice the size of the \textit{small} scenes, and the \textit{large} scenes are six times the size of the \textit{small} scenes, making them particularly challenging to handle.
  The Blender dataset is created using Blender and features varying ground textures, non-trivial geometries, and photorealistic rendering.
  The Blender dataset also contains the three scales.
  See the supplementary for details on data generation and view sampling.

  \textbf{Metric.}
  We render the inferred object-compositional NeRF into 2D RGB, depth and instance masks from each view.
  We report per-pixel root-mean-square-error (RMSE) on both RGB and depth value and compute mean-intersection-over-union (mIoU) score against ground truth masks.
  Pixels with depth values larger than the camera clipping depth are masked out. 
  We additionally report the L2 distance between inferred object location and the ground truth location.
  We report quantitative results for both input (I) and query (Q) views.

  \textbf{Baseline.}
  We compare our method with MulMON \citep{MulMON}, the state-of-the-art multi-view 3D scene object-centric learning method with online inference ability. 
  We adopt their official implementation and additionally add depth as both input and reconstruction targets.

  \subsection{Scalable Object Centric Learning}
  In practice, training scenes are typically of a limited scale, while the scenes to be deployed may have varying sizes. 
  To exam the test time scalability, we train the baseline and our model on small (MulMON\_small, Ours\_small) and medium (MulMON\_medium, Ours\_medium) scenes only and test them on all three scene scales.
  For our model, we fix $K$ to $7$ for all scene scales.
  For the baseline, we set the number of mixing components to be strictly larger than the total number of objects in the scenes during both training and testing.
  The quantitative results are reported in Table \ref{tab:S1_results}.
  The results of our approach are obtained as the average of $5$ runs.
  
  MulMON achieves $0.612$ mIoU when trained and tested on small scenes. 
  This setup aligns with their assumption that all objects appear in all views.
  However, when evaluated on medium and large-scale scenes, its performance drops to below $0.2$ mIoU.
  Training on medium scenes does not improve performance. 

\begin{figure}[h]
      \scriptsize
      \begin{center}
        \begin{tabular}{c@{\hskip1pt}c@{\hskip2pt}|@{\hskip2pt}c@{\hskip1pt}c@{\hskip2pt}|@{\hskip2pt}c@{\hskip1pt}c}
          
          \includegraphics[width=0.14\linewidth]{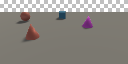} &
          \includegraphics[width=0.14\linewidth]{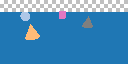} &
          \includegraphics[width=0.14\linewidth]{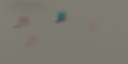} &
          \includegraphics[width=0.14\linewidth]{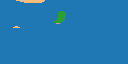} &
          \includegraphics[width=0.14\linewidth]{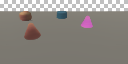} &
          \includegraphics[width=0.14\linewidth]{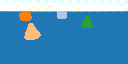} \\

          \includegraphics[width=0.14\linewidth]{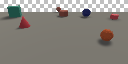} &
          \includegraphics[width=0.14\linewidth]{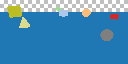} &
          \includegraphics[width=0.14\linewidth]{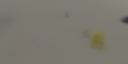} &
          \includegraphics[width=0.14\linewidth]{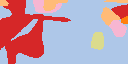} &
          \includegraphics[width=0.14\linewidth]{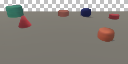} &
          \includegraphics[width=0.14\linewidth]{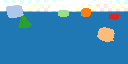} \\ \hline

          \multicolumn{2}{c|@{\hskip2pt}}{Ground Truth}&\multicolumn{2}{c|@{\hskip2pt}}{MulMON}&\multicolumn{2}{c}{Ours} \\ \hline

      \end{tabular}
      \end{center}
      \caption{Query view synthesis in large scenes with models trained on medium scenes.}
      \vspace{-2mm}
      \label{fig:S1_compare}
  \end{figure}

  By contrast, our model, being scene scale agnostic, can generalize to large scenes with a $0.07$ mIoU performance drop when trained on small scenes. 
  After training on medium scenes, our model generalizes to large scenes with no performance drop (see Fig. \ref{fig:S1_compare} for qualitative comparison). 
  The comparable performances on the input and query views demonstrate that our model infers view-invariant features resulting in robust rendering from any view.

  \begin{table*}[h]
      \scriptsize
      \centering
      \caption{Quantitative results on the Unity dataset.}
      \label{tab:S1_results}    
      \begin{tabular}{lc c@{\hskip6pt}c@{\hskip6pt}c c@{\hskip6pt}c@{\hskip6pt}c c@{\hskip6pt}c@{\hskip6pt}c c@{\hskip6pt}c@{\hskip6pt}c}                   
      \hline
      \multicolumn{2}{c}{}                   & \multicolumn{3}{c}{mIoU $\uparrow$}  & \multicolumn{3}{c}{RGB RMSE $\downarrow$} &  \multicolumn{3}{c}{depth RMSE $\downarrow$}  &  \multicolumn{3}{c}{L2 coord. error $\downarrow$ } \\ \cline{3-14}
      \multicolumn{2}{c}{}                   & s & m & l                            & s & m & l                                 &  s             & m      & l                                   &  s&m&l                                        \\ \hline
      \multirow{2}{*}{MulMON\_small \citep{MulMON}}  & I    & 0.612&0.198&0.158                    & \textbf{0.055}&0.167&0.178                &  \textbf{0.430}& 0.972  & 1.183                   &  \multicolumn{3}{c}{N/A}                      \\ \cline{2-14}
                                      & Q    & 0.599&0.192&0.152                    & 0.056&0.171&0.186                         &  0.455         & 1.001  & 1.140                   &  \multicolumn{3}{c}{N/A}                      \\ \hline
      \multirow{2}{*}{MulMON\_medium \citep{MulMON}} & I    & 0.371&0.225&0.141                    & 0.102&0.141&0.144                         &  0.891         & 0.991  & 1.223                  &  \multicolumn{3}{c}{N/A}                      \\ \cline{2-14}
                                      & Q    & 0.365&0.221&0.136                    & 0.103&0.149&0.159                         &  0.901         & 1.072  & 1.216                 &  \multicolumn{3}{c}{N/A}                      \\ \hline
      \multirow{2}{*}{Ours\_small}    & I    & \textbf{0.763}&0.721&0.694           & 0.074&0.107&0.141                         &  0.516&0.680&0.741                            &  \multicolumn{3}{c}{N/A}                      \\ \cline{2-14}
                                      & Q    & 0.761&0.713&0.690                    & 0.073&0.109&0.149                         &  0.517&0.691&0.723                            &  \textbf{0.068} & 0.170 & 0.192               \\ \hline
      \multirow{2}{*}{Ours\_medium}   & I    & 0.710&\textbf{0.756}&\textbf{0.761}  & 0.075&\textbf{0.098}&\textbf{0.091}       &  0.617&\textbf{0.650}&\textbf{0.634}          &  \multicolumn{3}{c}{N/A}                      \\ \cline{2-14}
                                      & Q    & 0.703&0.751&0.757                    & 0.073&0.099&0.092                         &  0.619&0.652&0.640                            &  0.099 & \textbf{0.117} & \textbf{0.111}      \\ \hline
    
      \end{tabular}                   
    \end{table*}

\begin{figure}[h]
    \begin{center}
    \scriptsize
        \begin{tabular}{@{}c@{\hskip1pt}|@{\hskip1pt}c@{\hskip1pt}c@{\hskip1pt}|@{\hskip1pt}c@{\hskip1pt}c@{\hskip1pt}c@{\hskip1pt}c@{\hskip1pt}c@{\hskip1pt}c@{}}
        
        \multirow{2}{*}[0.038\linewidth]{\includegraphics[width=0.145\linewidth]{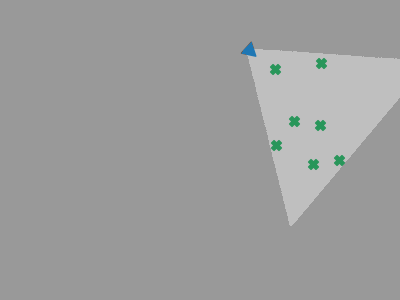}} 
                                    & \includegraphics[width=0.1\linewidth]{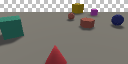} 
                                    & \includegraphics[width=0.1\linewidth]{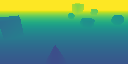} 
                                & \includegraphics[width=0.1\linewidth]{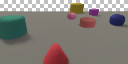} 
                                & \includegraphics[width=0.1\linewidth]{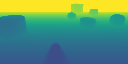} 
    
                                & \includegraphics[width=0.1\linewidth]{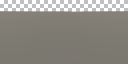}
                                & \includegraphics[width=0.1\linewidth]{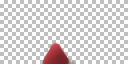}
                                & \includegraphics[width=0.1\linewidth]{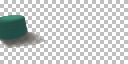}
                                & \includegraphics[width=0.1\linewidth]{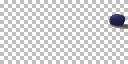}  \\
    
                                    & \includegraphics[width=0.1\linewidth]{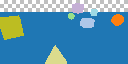} &                                                                 
                                & \includegraphics[width=0.1\linewidth]{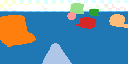} & 
                                    \includegraphics[width=0.1\linewidth]{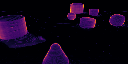}   
    
                                    & \includegraphics[width=0.1\linewidth]{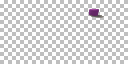}
                                    & \includegraphics[width=0.1\linewidth]{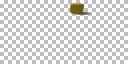}
                                    & \includegraphics[width=0.1\linewidth]{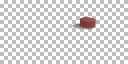}
                                    & \includegraphics[width=0.1\linewidth]{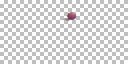}  \\
    
        \multirow{2}{*}[0.038\linewidth]{\includegraphics[width=0.145\linewidth]{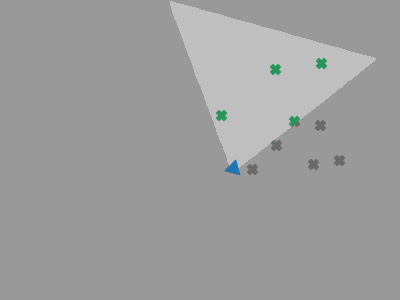}} 
                                    & \includegraphics[width=0.1\linewidth]{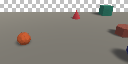} 
                                    & \includegraphics[width=0.1\linewidth]{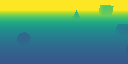} 
                                & \includegraphics[width=0.1\linewidth]{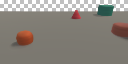} 
                                & \includegraphics[width=0.1\linewidth]{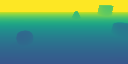} 
    
                                & \includegraphics[width=0.1\linewidth]{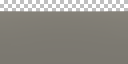}
                                & \includegraphics[width=0.1\linewidth]{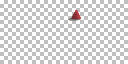}
                                & \includegraphics[width=0.1\linewidth]{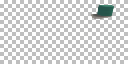}
                                & \includegraphics[width=0.1\linewidth]{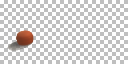}  \\
    
                                    & \includegraphics[width=0.1\linewidth]{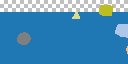} &                                                                 
                                & \includegraphics[width=0.1\linewidth]{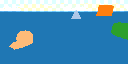} & 
                                    \includegraphics[width=0.1\linewidth]{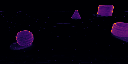}   
    
                                    & \includegraphics[width=0.1\linewidth]{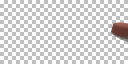}
                                    & \includegraphics[width=0.1\linewidth]{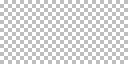}
                                    & \includegraphics[width=0.1\linewidth]{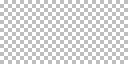}
                                    & \includegraphics[width=0.1\linewidth]{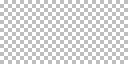}  \\
    
        \multirow{2}{*}[0.038\linewidth]{\includegraphics[width=0.145\linewidth]{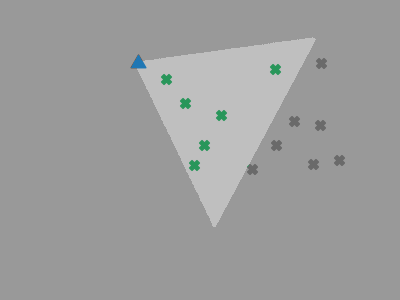}} 
                                    & \includegraphics[width=0.1\linewidth]{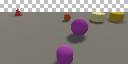} 
                                    & \includegraphics[width=0.1\linewidth]{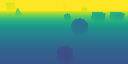} 
                                & \includegraphics[width=0.1\linewidth]{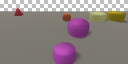} 
                                & \includegraphics[width=0.1\linewidth]{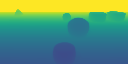} 
    
                                & \includegraphics[width=0.1\linewidth]{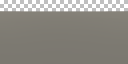}
                                & \includegraphics[width=0.1\linewidth]{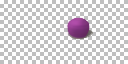}
                                & \includegraphics[width=0.1\linewidth]{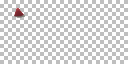}
                                & \includegraphics[width=0.1\linewidth]{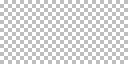}  \\
    
                                    & \includegraphics[width=0.1\linewidth]{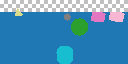} &                                                                 
                                & \includegraphics[width=0.1\linewidth]{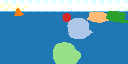} & 
                                    \includegraphics[width=0.1\linewidth]{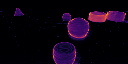}   
    
                                    & \includegraphics[width=0.1\linewidth]{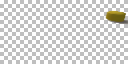}
                                    & \includegraphics[width=0.1\linewidth]{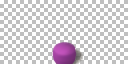}
                                    & \includegraphics[width=0.1\linewidth]{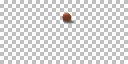}
                                    & \includegraphics[width=0.1\linewidth]{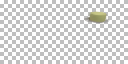}  \\
    
        \multirow{2}{*}[0.038\linewidth]{\includegraphics[width=0.145\linewidth]{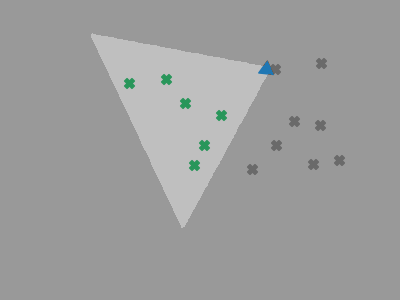}} 
                                    & \includegraphics[width=0.1\linewidth]{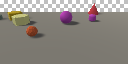} 
                                    & \includegraphics[width=0.1\linewidth]{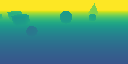} 
                                & \includegraphics[width=0.1\linewidth]{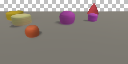} 
                                & \includegraphics[width=0.1\linewidth]{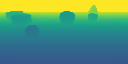} 
    
                                & \includegraphics[width=0.1\linewidth]{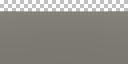}
                                & \includegraphics[width=0.1\linewidth]{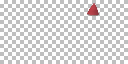}
                                & \includegraphics[width=0.1\linewidth]{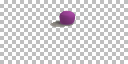}
                                & \includegraphics[width=0.1\linewidth]{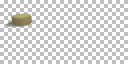}  \\
    
                                    & \includegraphics[width=0.1\linewidth]{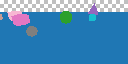} &                                                                 
                                & \includegraphics[width=0.1\linewidth]{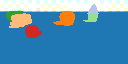} & 
                                    \includegraphics[width=0.1\linewidth]{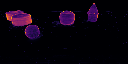}   
    
                                    & \includegraphics[width=0.1\linewidth]{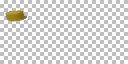}
                                    & \includegraphics[width=0.1\linewidth]{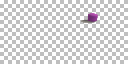}
                                    & \includegraphics[width=0.1\linewidth]{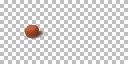}
                                    & \includegraphics[width=0.1\linewidth]{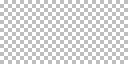}  \\ \hline

    Cog. Map & \multicolumn{2}{c|@{\hskip1pt}}{Ground Truth} & \multicolumn{6}{c}{Reconstruction} \\ \hline
    
        \end{tabular}
    \end{center}
    \caption{An online inference process from top to bottom.
    Each row corresponds to one update step. 
    The left column shows the evolution of the cognitive map.
    The camera pose of each step is marked by a blue triangle and the camera cone (FOV) is highlighted.
    Each object latent registered in the cognitive map is marked with an x and is greyed out if outside of view.}
    \label{fig:S1}
\end{figure}

  In Fig. \ref{fig:S1} we visualize a 4-step online inference process. 
  For each step, our model discovers new objects, updates the corresponding latent variables and registers them to the cognitive map.
  The inference process can repeat indefinitely to cover arbitrarily large areas.
  In the supplementary, we present additional qualitative results and further experiments exploring various aspects of our model.

  To examine the ability to handle non-trivial geometries and higher visual complexity, we conduct experiments on the Blender dataset.
  We additionally apply per-object finetuning on top of the inference results.
  Quantitative results are reported in Table \ref{tab:S2_results}.

\begin{figure}[h]
      \begin{center}
      \scriptsize
        \begin{tabular}{@{}c@{\hskip1pt}|@{\hskip1pt}c@{\hskip1pt}|@{\hskip1pt}c@{\hskip1pt}|@{\hskip1pt}c@{\hskip1pt}|@{\hskip1pt}c@{\hskip1pt}c@{}}
    
                \includegraphics[width=0.15\linewidth]{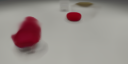} 
              & \includegraphics[width=0.15\linewidth]{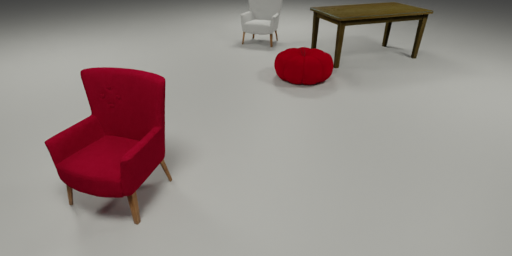} 
              & \includegraphics[width=0.15\linewidth]{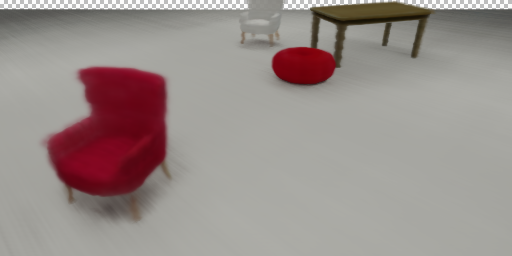} 
              & \includegraphics[width=0.15\linewidth]{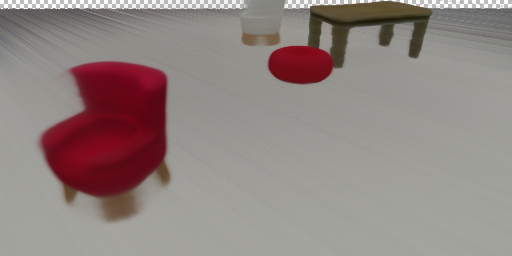} 
              & \includegraphics[width=0.15\linewidth]{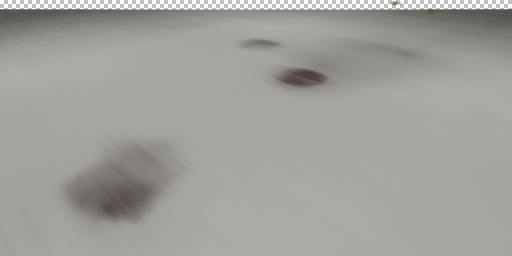} 
              & \includegraphics[width=0.15\linewidth]{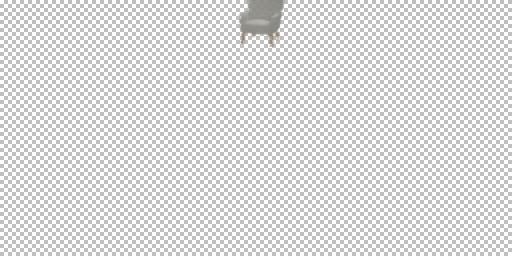}  \\
    
                \includegraphics[width=0.15\linewidth]{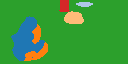} 
              & \includegraphics[width=0.15\linewidth]{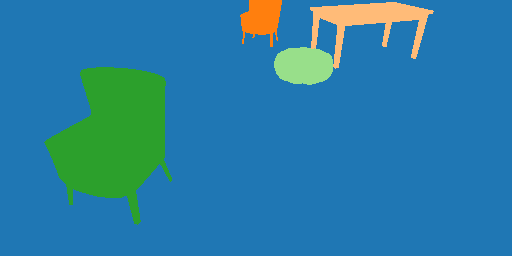} 
              & \includegraphics[width=0.15\linewidth]{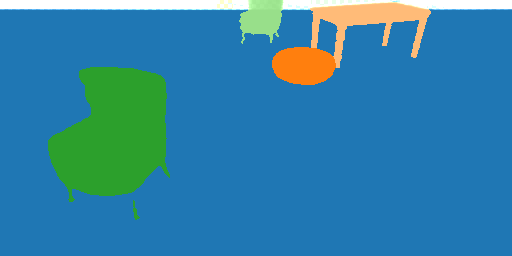} 
              & \includegraphics[width=0.15\linewidth]{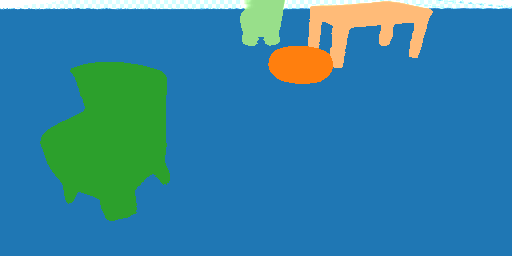} 
              & \includegraphics[width=0.15\linewidth]{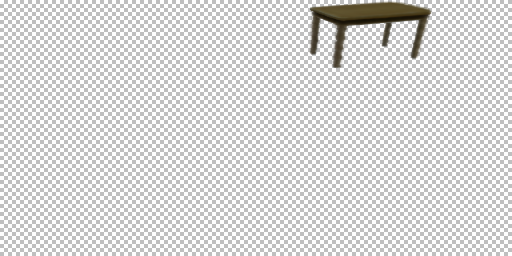} 
              & \includegraphics[width=0.15\linewidth]{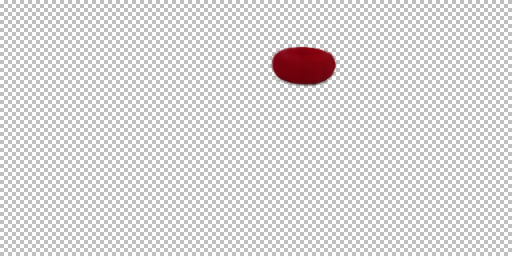}   \\ 
              
              \includegraphics[width=0.15\linewidth]{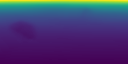} 
              & \includegraphics[width=0.15\linewidth]{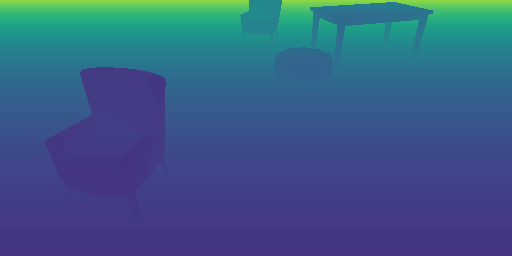} 
              & \includegraphics[width=0.15\linewidth]{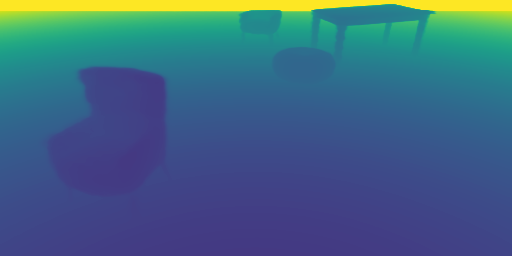} 
              & \includegraphics[width=0.15\linewidth]{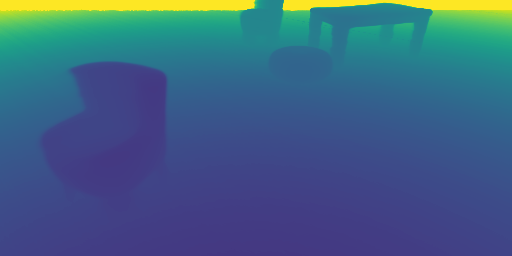} 
              & \includegraphics[width=0.15\linewidth]{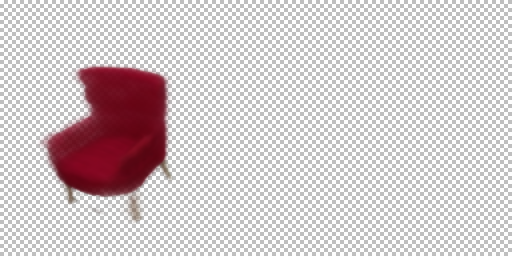} 
              & \includegraphics[width=0.15\linewidth]{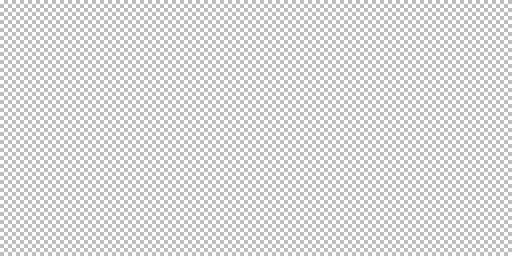}   \\ 
                  
              \includegraphics[width=0.15\linewidth]{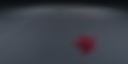} 
              & \includegraphics[width=0.15\linewidth]{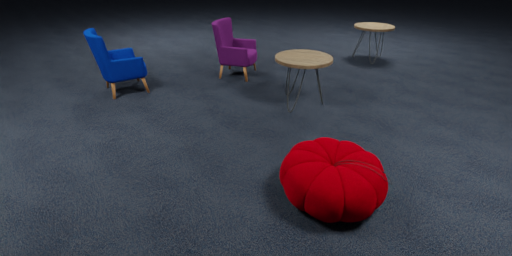} 
              & \includegraphics[width=0.15\linewidth]{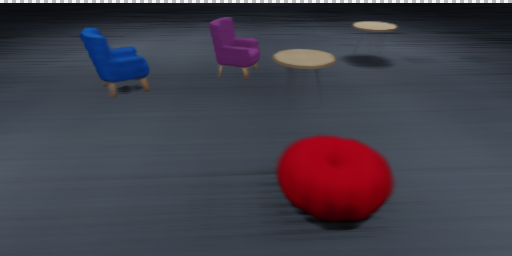} 
              & \includegraphics[width=0.15\linewidth]{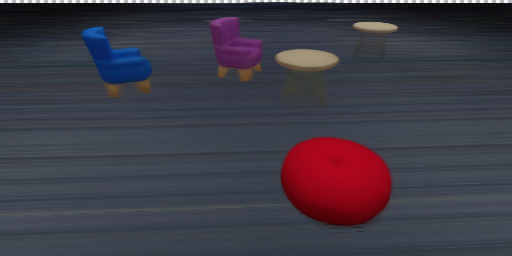} 
              & \includegraphics[width=0.15\linewidth]{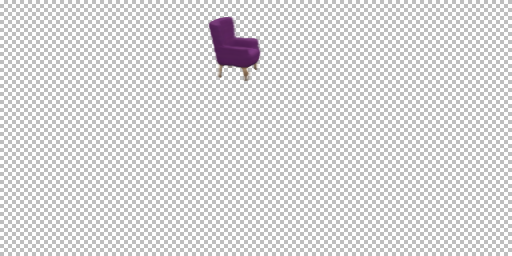} 
              & \includegraphics[width=0.15\linewidth]{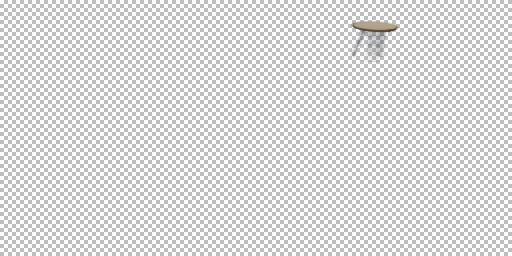}  \\
    
                \includegraphics[width=0.15\linewidth]{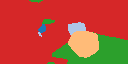} 
              & \includegraphics[width=0.15\linewidth]{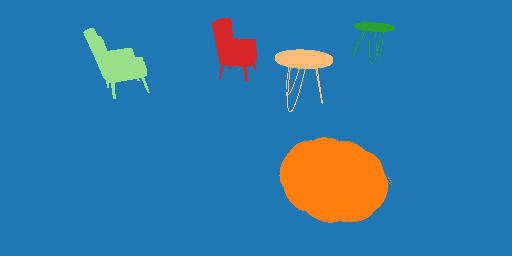} 
              & \includegraphics[width=0.15\linewidth]{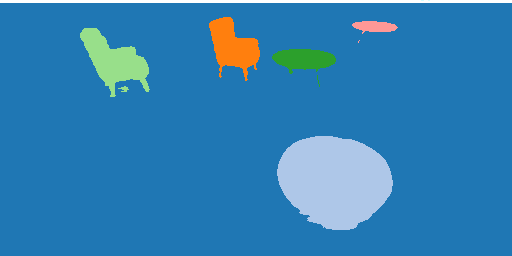} 
              & \includegraphics[width=0.15\linewidth]{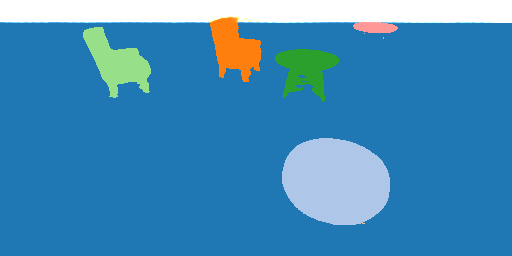} 
              & \includegraphics[width=0.15\linewidth]{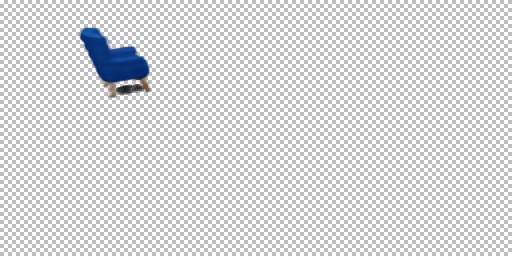} 
              & \includegraphics[width=0.15\linewidth]{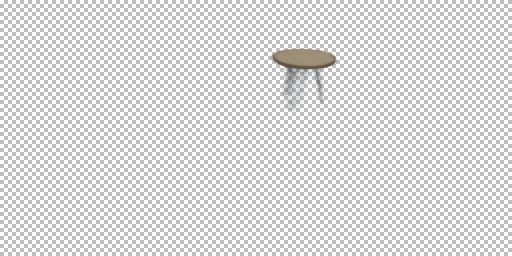}   \\ 
              
                \includegraphics[width=0.15\linewidth]{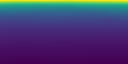} 
              & \includegraphics[width=0.15\linewidth]{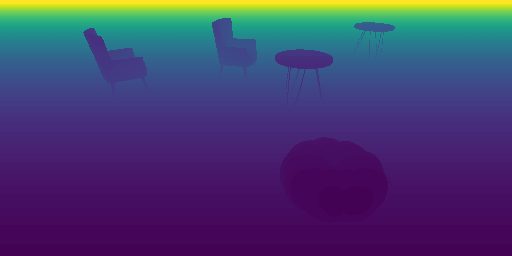} 
              & \includegraphics[width=0.15\linewidth]{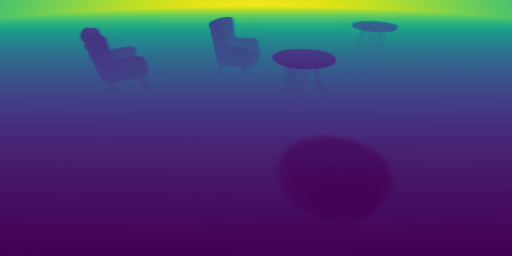} 
              & \includegraphics[width=0.15\linewidth]{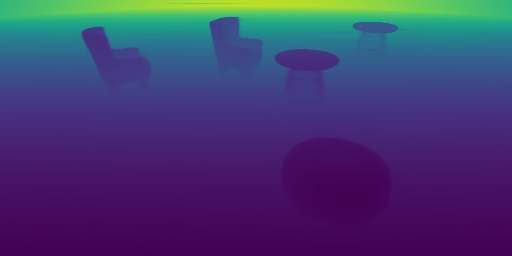} 
              & \includegraphics[width=0.15\linewidth]{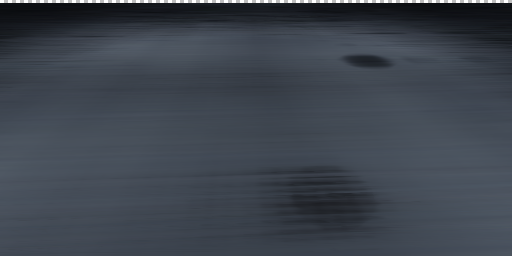} 
              & \includegraphics[width=0.15\linewidth]{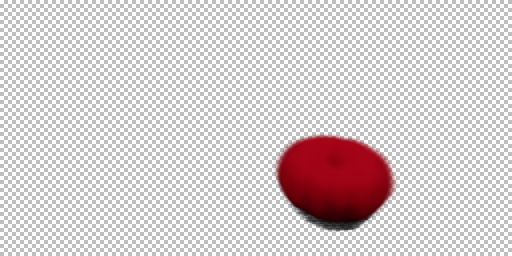}   \\ \hline

      MulMON & GT & SOOC3D+ & SOOC3D &\multicolumn{2}{c}{SOOC3D+ Components} \\ \hline
    
      \end{tabular}
      \end{center}
      \caption{\textbf{Small scenes} (top three rows) and \textbf{large scenes} (bottom three rows) novel view synthesis produced by the baseline (MulMON), our model (SOOC3D), and per-object finetuning (SOOC3D+).}
      \label{fig:b_compare}
    \end{figure}

  When dealing with complex geometries, our proposed model infers accurate object structures and outperforms the baseline in the small scene setup, while the baseline model tends to predict bubble-like shapes (Fig. \ref{fig:b_compare}). 
  Additionally, after training on medium-sized scenes, our model generalizes to large scenes without any performance degradation, being scene scale agnostic. 
  In contrast, the performance of the baseline model significantly deteriorates in large scenes (Fig. \ref{fig:b_compare}). 


  We apply per-object finetuning to the Blender dataset inference results. 
  As reported in Table \ref{tab:S2_results} last row, the performance increases across all metrics.
  Object structure details are recovered with preserved object identities.
  Finetuned NeRFs can be registered into the cognitive for future queries.

    \begin{table*}[h]
    \scriptsize
    \centering
    \caption{Quantitative results on the Blender dataset.}
    \label{tab:S2_results}    
    \begin{tabular}{lc c@{\hskip6pt}c@{\hskip6pt}c c@{\hskip6pt}c@{\hskip6pt}c c@{\hskip6pt}c@{\hskip6pt}c}                   
    \hline
    \multicolumn{2}{c}{}                   & \multicolumn{3}{c}{mIoU $\uparrow$}              & \multicolumn{3}{c}{RGB RMSE $\downarrow$}        & \multicolumn{3}{c}{depth RMSE $\downarrow$}                            \\ \cline{3-11}
    \multicolumn{2}{c}{}                   & s     &  m    &  l                               &  s    & m     & l                                 & s&m&l                                                                  \\ \hline
    \multirow{2}{*}{MulMON\_small \citep{MulMON}}  & I    & 0.492 & 0.182 & 0.121             & 0.070 & 0.131 & 0.139                             & 0.422 & 0.438 & 0.451                                                  \\ \cline{2-11}
                                    & Q    & 0.489 & 0.176 & 0.114                            & 0.069 & 0.132 & 0.138                             & 0.423 & 0.440 & 0.452                                                  \\ \hline
    \multirow{2}{*}{MulMON\_medium \citep{MulMON}} & I    & 0.311 & 0.185 & 0.106             & 0.124 & 0.127 & 0.136                             & 0.432 & 0.454 & 0.450                                                  \\ \cline{2-11}
                                    & Q    & 0.304 & 0.171 & 0.100                            & 0.122 & 0.129 & 0.137                             & 0.435 & 0.446 & 0.453                                                  \\ \hline
    \multirow{2}{*}{Ours\_small}    & I    & 0.741 & 0.667 & 0.661                            & 0.071 & 0.084 & 0.086                             & 0.402 & 0.413 & 0.419                                                  \\ \cline{2-11}
                                    & Q    & 0.734 & 0.668 & 0.659                            & 0.072 & 0.089 & 0.090                             & 0.401 & 0.415 & 0.418                                                  \\ \hline
    \multirow{2}{*}{Ours\_medium}   & I    & 0.699 & 0.684 & 0.682                            & 0.088 & 0.086 & 0.082                             & 0.409 & 0.411 & 0.414                                                  \\ \cline{2-11}
                                    & Q    & 0.698 & 0.673 & 0.678                            & 0.085 & 0.083 & 0.084                             & 0.407 & 0.412 & 0.416                                                  \\ \hline
    \multirow{2}{*}{Ours\_finetune} & I    & \textbf{0.861} & \textbf{0.848} & \textbf{0.830} & \textbf{0.026} & \textbf{0.025} & \textbf{0.023} & \textbf{0.248} & \textbf{0.251} & \textbf{0.259}                       \\ \cline{2-11}
                                    & Q    & 0.858 & 0.839 & 0.826                            & 0.031 & 0.029 & 0.028                            & 0.250 & 0.254 & 0.264                                                  \\ \hline
    \end{tabular}                  
  \end{table*}
  
    \begin{figure}[H]
      \scriptsize
      \begin{center}
        \begin{tabular}{c@{\hskip1pt}c@{\hskip1pt}c@{\hskip1pt}c@{\hskip1pt}c@{\hskip1pt}}
    
          \includegraphics[width=0.19\linewidth]{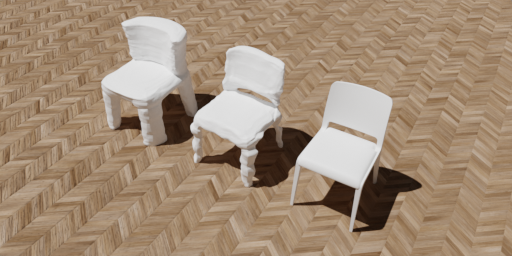}                    &
          \includegraphics[width=0.19\linewidth]{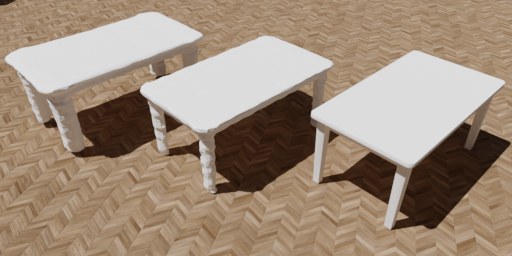}                    & 
          \includegraphics[width=0.19\linewidth]{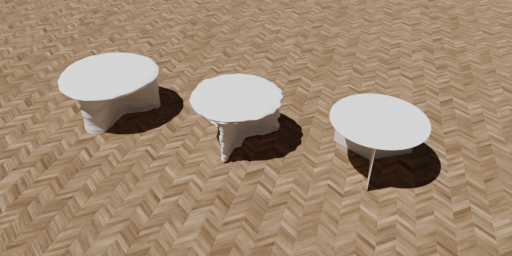}        &
    
          \includegraphics[width=0.19\linewidth]{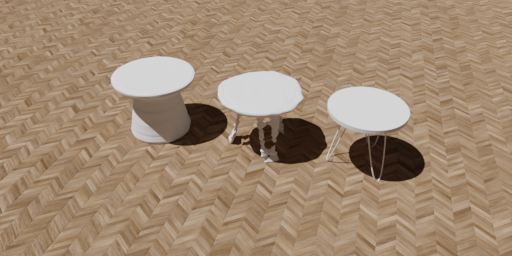}              &
          \includegraphics[width=0.19\linewidth]{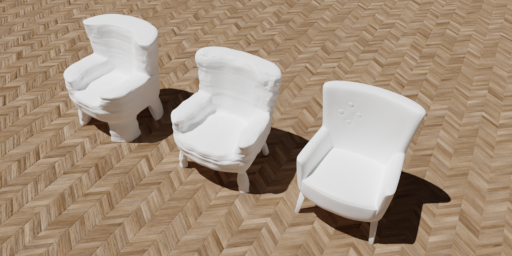}                  \\
    
        \end{tabular}
      \end{center}
      \caption{Mesh reconstruction visualization. In each image from left to right is the SOOC3D inference result, SOOC3D+ (per-object finetuning) result, and ground truth mesh.}
      \label{fig:mesh}
  \end{figure}

    To evaluate the 3D object reconstruction quality, we extract meshes from NeRF components via marching cube and compute accuracy and completeness scores against ground truth meshes as defined in previous works \cite{MeshRec}.
    We report that without finetuning, our model achieves on average 7.8 cm accuracy and 5.4 cm completeness. 
    After finetuning, we achieve on average 5.3 cm accuracy and 4.2 cm completeness. 
    Note that this reconstruction result is obtained by only observing limited sparse views.
    As shown by the qualitative comparison in Fig. \ref{fig:mesh}, the per-object finetuning process recovers fine details including extremely thin structures.

  \subsection{Real-world Dataset Results}

  \begin{wraptable}{r}{4cm}
    \caption{HM3D segmentation results.}\label{tab:hm3d}
    \begin{tabular}{cc}  \\  \toprule  
            & mIoU       \\  \midrule
    MulMON \citep{MulMON}  &  0.134     \\  \midrule
    SOOC3D  &  0.396     \\  \midrule
    SOOC3D+ &  0.521     \\  \bottomrule
    \end{tabular}
    \end{wraptable} 

  To demonstrate the potential of our method, we apply our model to Habitat-Matterport 3D (HM3D) \cite{HMP3D}, a real-world dataset comprised of textured meshes obtained from scanned indoor scenes.
  In our experiment, we specifically focus on segmenting salient furniture with regular structures, such as chairs, tables, and sofas.
  We thus render RGBD data from a set of 50 scenes within the dataset.
  The object segmentation performance is reported in Table \ref{tab:hm3d}.
The baseline model~\citep{MulMON} struggled in inferring correct object identities due to the high visual complexity of scenes.

\begin{figure}[h]
    \begin{center}
    \scriptsize
      \begin{tabular}{@{}c@{\hskip1pt}@{\hskip1pt}c@{\hskip1pt}@{\hskip1pt}c@{\hskip1pt} | @{\hskip1pt}c@{\hskip1pt}@{\hskip1pt}c@{\hskip1pt}c@{}}
  
              \includegraphics[width=0.15\linewidth]{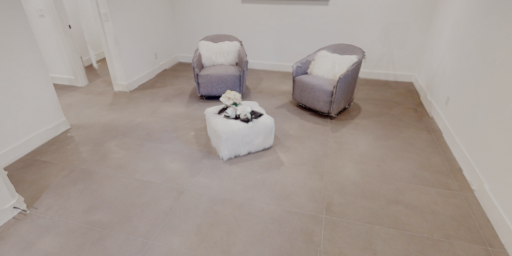} 
            & \includegraphics[width=0.15\linewidth]{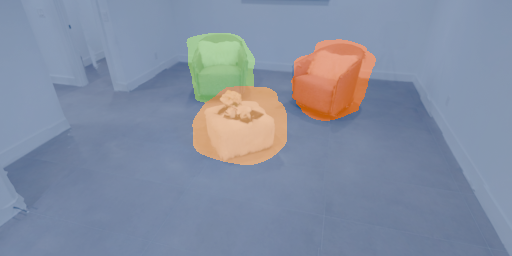} 
            & \includegraphics[width=0.15\linewidth]{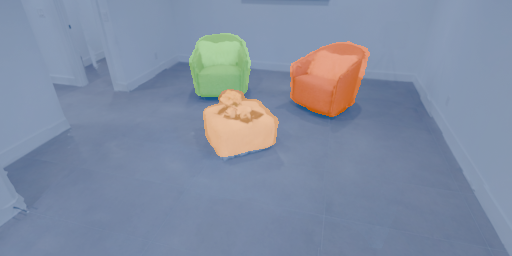} 
            & \includegraphics[width=0.15\linewidth]{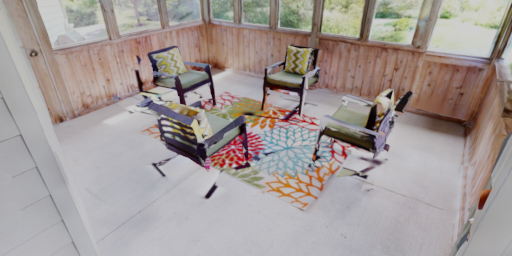} 
            & \includegraphics[width=0.15\linewidth]{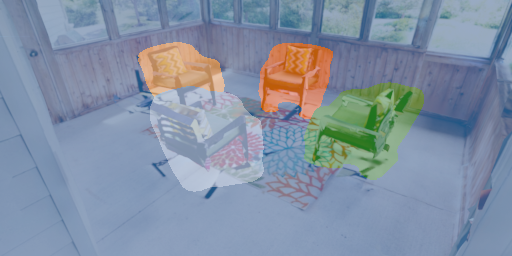} 
            & \includegraphics[width=0.15\linewidth]{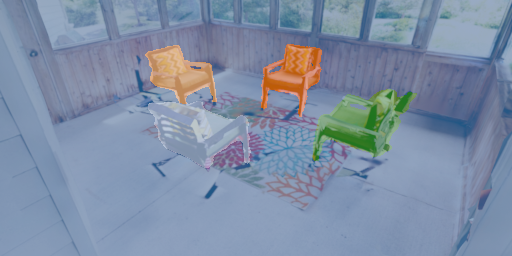}  \\
  
              \includegraphics[width=0.15\linewidth]{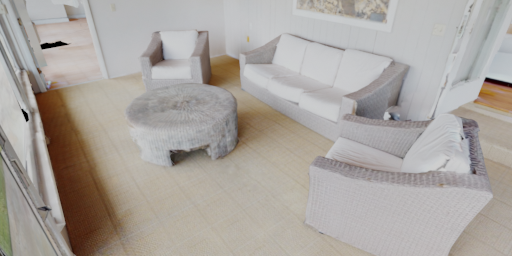} 
            & \includegraphics[width=0.15\linewidth]{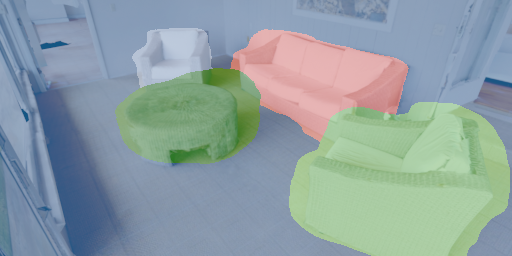} 
            & \includegraphics[width=0.15\linewidth]{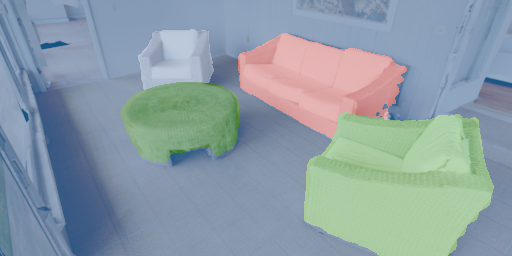} 
            & \includegraphics[width=0.15\linewidth]{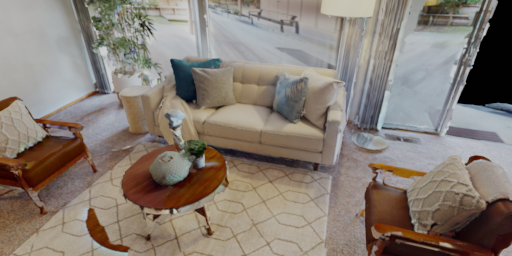} 
            & \includegraphics[width=0.15\linewidth]{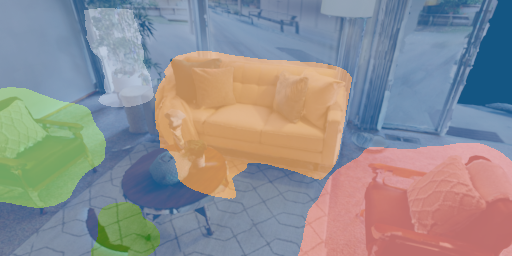} 
            & \includegraphics[width=0.15\linewidth]{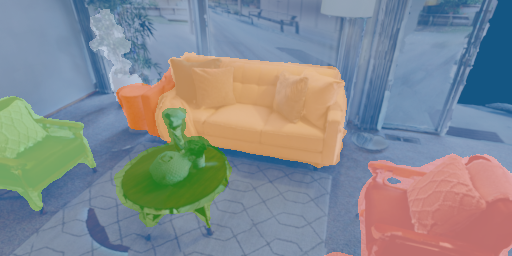}  \\

              \includegraphics[width=0.15\linewidth]{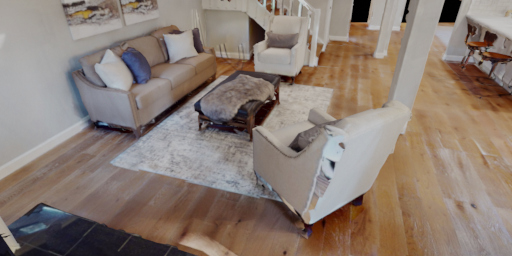} 
            & \includegraphics[width=0.15\linewidth]{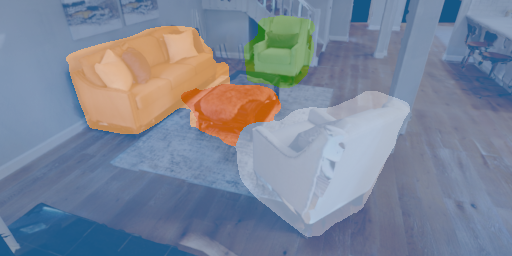} 
            & \includegraphics[width=0.15\linewidth]{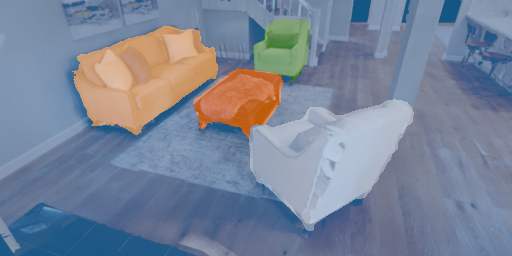} 
            & \includegraphics[width=0.15\linewidth]{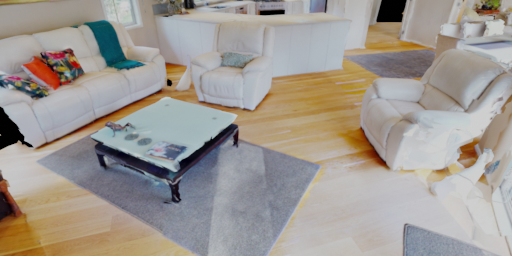}   
            & \includegraphics[width=0.15\linewidth]{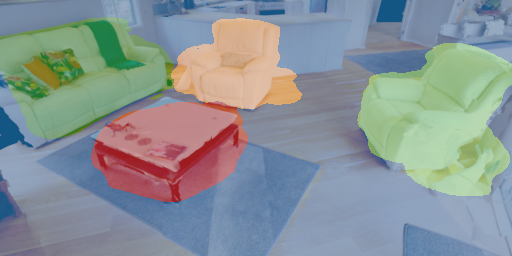}  
            & \includegraphics[width=0.15\linewidth]{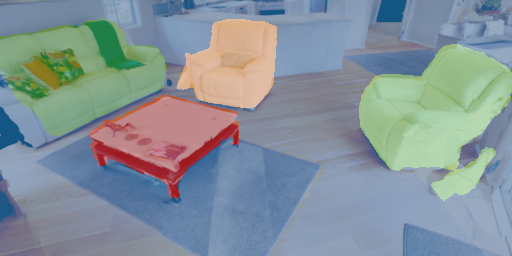}  \\ \hline

    Observation & SOOC3D & SOOC3D+ & Observation & SOOC3D & SOOC3D+ \\ \hline
  
    \end{tabular}
    \end{center}
    \caption{Predicted furniture instance masks on HM3D dataset.}
    \label{fig:hm3d}
  \end{figure}

  As shown in Fig. \ref{fig:hm3d}, our model learns to segment furniture and infer their basic structures.
  The per-object finetuning process significantly improves the segmentation quality with preserved object identities.
  We do observe that our model tends to overlook small-scale objects such as sofa cushions or floral displays on tables.

  \section{Limitation}
  Our pipeline aims to recover 3D scene geometry and enable scalable inference.
  However, there are two limitations to consider.
  First, we only model 3 DoF for object poses instead of 6.
  As a result, the appearance embedding of an object in a standing position differs from that of the same object in a lying down position.
  Second, our model is designed for handling a static scene and is not suitable for handling moving objects. 
  To overcome these limitations, one can use the same variational inference formulation and incorporate a transition model to predict 6 DoF object motions between time steps.
  
  \section{Conclusion and Future Work}
  We propose a framework for unsupervised 3D object-centric learning for handling scenes of large scale and a varying number of objects in the scene. 
  We introduced factorized latent learning which separates the object pose and view-invariant appearance latent variables.
  Our object-compositional NeRF allows the learning of 3D representation in the object coordinate system. 
  The cognitive map ensures object permanence and keeps track of all detected objects. 
  The inference results on HM3D demonstrate the potential of 3D object-centric learning algorithms.
  Our learned view-invariant 3D object representation can potentially be applied in the SLAM system or relational reasoning tasks. 
  In future work, we aim to achieve 6 DoF object pose estimation and dynamic scene modeling. 

\clearpage

\bibliography{icml2023}
\bibliographystyle{unsrtnat}


\end{document}